\documentclass[10pt,onecolumn,letterpaper]{article}

\usepackage[T1]{fontenc}
\usepackage{lmodern}
\usepackage[margin=0.75in]{geometry}
\usepackage[numbers,sort&compress]{natbib}
\usepackage{xcolor}

\usepackage{etoolbox}

\usepackage{amsmath,amssymb}

\DeclareMathOperator*{\argmax}{arg\,max}

\usepackage{booktabs}
\usepackage{multirow}
\usepackage{graphicx}
\usepackage{subcaption}
\usepackage{float}  

\usepackage{algorithm}
\usepackage{algorithmic}

\usepackage{pifont}  

\usepackage{xspace}
\newcommand{\method}{IVT\xspace}  
\newcommand{\methodfull}{Iterative Visual Thinking\xspace}

\makeatletter
\DeclareRobustCommand\onedot{\futurelet\@let@token\@onedot}
\def\@onedot{\ifx\@let@token.\else.\null\fi\xspace}
\makeatother
\def\eg{\emph{e.g}\onedot} 
\def\ie{\emph{i.e}\onedot}

\def\etal{\emph{et al}\onedot}

\providecommand{\linenumbers}{}
\providecommand{\nolinenumbers}{}
\AtBeginEnvironment{enumerate}{\nolinenumbers}
\AtEndEnvironment{enumerate}{\linenumbers}
\AtBeginEnvironment{itemize}{\nolinenumbers}
\AtEndEnvironment{itemize}{\linenumbers}


\definecolor{linkblue}{rgb}{0.21,0.49,0.74}
\usepackage[pagebackref,breaklinks,colorlinks,allcolors=linkblue]{hyperref}
\usepackage[capitalize]{cleveref}

\title{Iterative Visual Thinking and the\\Self-Correction Mirage in VLM Grounding}

\author{Animesh Tripathy\\
QpiAI India Pvt. Ltd.
\and
Aswanth Krishnan\\
QpiAI India Pvt. Ltd.
}
\date{}

\begin{document}
\maketitle

\begin{abstract}
Letting a vision-language model (VLM) think longer at test time has driven much
recent progress. A natural way to bring this to spatial grounding is \emph{visual
self-correction}: the model predicts a bounding box, sees it rendered on the image,
and refines it over several steps. We build a faithful instance of this
idea, \methodfull{} (\method{}), with a two-phase recipe: a supervised warm-up in
which the base model's own predictions serve as realistic errors that a teacher VLM
turns into corrective reasoning traces (yielding training data without human
annotation), followed by GRPO with a simple IoU reward. Measured the way such
systems are usually reported, it works: the trained model surpasses the single-shot
base by $+2.4$pp Acc@0.5.

We show this gain is a \textbf{measurement mirage}. The reported number silently
keeps, per sample, the trajectory step closest to the ground-truth box: an oracle
that needs the very answer it predicts. Re-scored under deployable, label-free
stopping rules the improvement vanishes, and the best policy is not to iterate at
all: stopping at step~0 matches the base and beats every shippable rule. The cause
is a \textbf{verification failure}, since the model can generate a better box
somewhere in its trajectory but cannot identify it. Self-verification confidence
correlates only weakly with correctness ($r\!\approx\!0.22$), and a counterfactual
overlay shows the loop reacts to the \emph{presence} of a rendered box rather than its
correctness. We distill the lesson into an \textbf{honest-trajectory evaluation
protocol}, accuracy under fixed label-free policies plus an explicit
oracle--shippable gap.
\end{abstract}

\section{Introduction}
\label{sec:intro}

Scaling test-time computation has become one of the most reliable ways to improve
reasoning: systems such as o1 and DeepSeek-R1~\cite{deepseekr1} ``think'' for longer
before answering, and a growing line of work extends this slow-thinking approach to
vision by letting models reason \emph{through} images: zooming, cropping, or
manipulating visual representations~\cite{qi2025cogcom,fan2025grit,sarch2025vigorl}. A
natural and appealing instance for spatial grounding is \emph{visual self-correction}:
given a referring expression such as ``the back of a zebra looking to the left,'' a
vision-language model (VLM) predicts a bounding box, \emph{sees} that box rendered as an
overlay on the image, and iteratively refines it (\cref{fig:loop}). This mirrors how
people localize objects (look, hypothesize, check, correct) and promises a closed-loop,
interpretable alternative to single-shot prediction.

This paper asks a deceptively simple question: \textbf{does test-time visual
self-correction actually help VLM grounding?} The answer, on a strong open model
and a challenging multi-dataset benchmark, is \emph{no}: not in any way that can be
deployed. We reach this conclusion not by failing to make the method work, but by auditing
a pipeline that, by the usual reporting conventions, \emph{does} work.\footnote{An earlier
version of this work reported only best-step accuracies and concluded that the iterative
method beats the base model. Here we additionally report deployment-aware, label-free
policies and find that the gain does not survive them. The method, the training recipe, and
the measured trajectories are identical; only the evaluation, and therefore the
conclusion, has changed.}

\paragraph{A method that appears to succeed.}
We study \methodfull{} (\method{}): predict, render, refine, trained with a two-phase
recipe we describe in \cref{sec:method}. Consistent with prior reports that VLMs cannot
self-correct without training~\cite{huangcorrect,liao2025can}, na\"ively prompting a strong
grounder (Qwen3-VL-4B, $79.6\%$ Acc@0.5 single-shot) to iterate over its rendered
predictions \emph{collapses} accuracy. A two-phase recipe, supervised warm-up on
teacher-generated correction traces followed by GRPO~\cite{deepseekmath,deepseekr1} with a
simple IoU reward, repairs the collapse and, on paper, \emph{surpasses} the base model: Acc@0.5
rises to $82.0\%$ ($+2.4$pp), with gains at the stricter thresholds too. Reported this way,
slow thinking looks like a clear win for grounding.

\paragraph{The win is an oracle.}
The gain disappears under scrutiny. The reported $82.0\%$ is a \emph{best-step-per-sample}
number: the evaluation silently keeps, for each test image, the step in the trajectory whose
box best matches the ground truth. This is an \textbf{oracle} that cannot exist at inference,
because choosing the best step requires the very answer we are trying to predict. When we
instead apply policies a deployed system could actually run (always take the last step, stop
when the box stabilizes, stop when motion is small, keep the last valid box), the improvement
vanishes (\cref{sec:mirage}). The single best policy turns out to be \emph{not to iterate at
all}: stopping at step~0 matches the base model and beats every shippable stopping rule. The
artifact is not subtle once named: it even inflates the headline ``collapse.'' What is usually
reported as a $31$pp drop for na\"ive iteration ($79.6\!\rightarrow\!48.7\%$) is itself an
oracle/last-valid figure; the actual, last-step accuracy a user would see is $17.0\%$, a
$63$pp collapse. GRPO, credited in the original framing with ``stabilizing refinement,''
turns out to be purely \emph{defensive}: it shrinks per-step degradation $5\times$ but never
recovers a gain over not iterating.

\paragraph{Why no observed signal fixes it: the verification gap.}
A practical adaptive policy could in principle recover the oracle if it knew \emph{when} to
stop. None of the signals we test supplies that knowledge. The oracle's gain is tiny and concentrated in
the $\sim$20\% hardest cases (where iteration nudges mean IoU from $0.10$ to $0.20$, \ie,
from one kind of failure to another), while on the $66\%$ easy cases iterating only hurts
(\cref{sec:mirage}). No observable signal predicts which hard case will benefit: expression
length is essentially uncorrelated with whether a later step helps ($r\!\approx\!-0.14$), and
box geometry is uninformative. We then test the most direct signal, \emph{self-verification}:
render each step and ask the model ``is this box correct?''~\cite{kadavath2022know}.
Confidence correlates only weakly with actual correctness (Pearson $r\!\approx\!0.22$), the
model is badly miscalibrated (mean $P(\text{yes})\!\approx\!0.40$ even on correct boxes), and
using it to gate stopping \emph{still} loses to not iterating (\cref{sec:verification}). A
counterfactual overlay sharpens the diagnosis: the loop reacts to the \emph{presence} of a
rendered box, not its \emph{content}. Painting the ground-truth box on the image barely helps,
and a mirrored wrong box is not tracked. The model can refine, but it cannot tell which of its
own attempts is right, so slow thinking cannot be cashed out into shippable accuracy.

\paragraph{Contributions.}
\begin{itemize}
  \item A critical study of test-time visual self-correction for grounding, showing that a
  recipe which appears to beat the base model ($+2.4$pp) provides \emph{no deployable gain}
  once oracle best-step selection is removed, and that the true na\"ive collapse
  ($-63$pp) is larger than the oracle-softened figure usually reported (\cref{sec:mirage}).
  \item Evidence that the missing ingredient is \emph{verification}: the oracle gain is small,
  concentrated in cases that remain failures, and unpredictable from any observable signal we
  test, including the model's own self-verification confidence (weakly correlated with
  correctness, miscalibrated) and a counterfactual overlay that hands the model the answer
  (\cref{sec:verification}).
  \item A clarification of GRPO's role in iterative grounding: it is \emph{defensive} (halves
  per-step degradation) rather than additive, never beating a no-iteration baseline.
  \item An \textbf{honest-trajectory evaluation protocol} (\cref{sec:protocol}): report
  accuracy under fixed, label-free stopping policies together with an explicit
  oracle--shippable gap and an attainable-gain term, so iterative ``gains'' are credited only
  when reachable in deployment.
\end{itemize}

\begin{figure*}[t]
  \centering
  \includegraphics[width=0.82\linewidth]{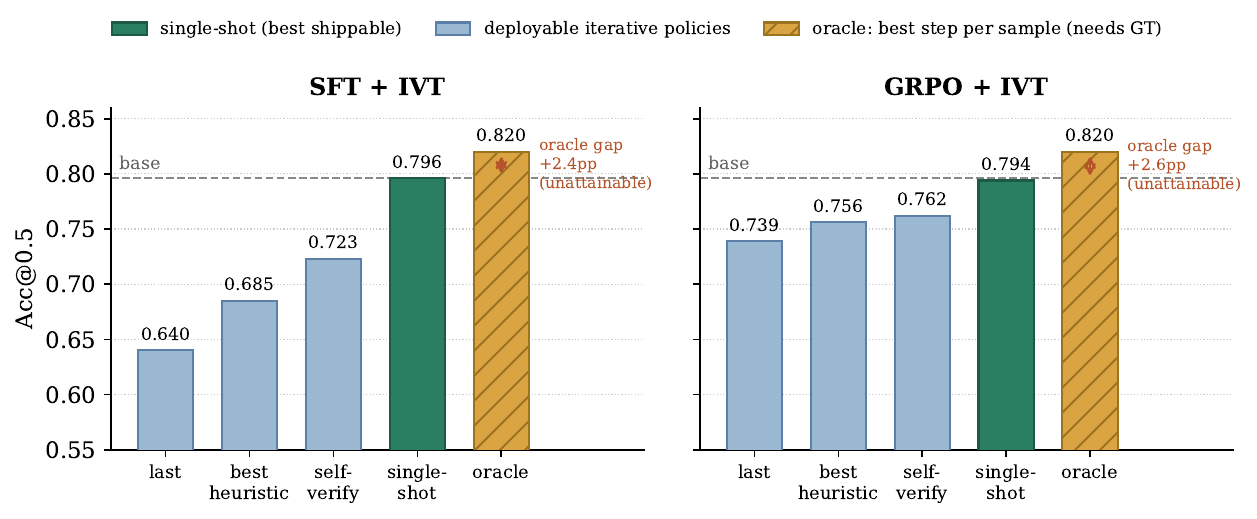}
  \caption{\textbf{The self-correction mirage (Acc@0.5, $505$-sample test set).} The standard
  \emph{oracle} report keeps the best step per sample and so requires the ground-truth box
  (gold, hatched); it shows a $+2.4$pp gain over the base model. But every \emph{deployable}
  policy, whether taking the last step, the best box-geometry heuristic, or the step the model
  self-verifies as correct, falls \emph{below} not iterating at all (\emph{single-shot}),
  which simply matches the base model (dashed line). The ``self-correction'' gain is an oracle
  artifact: the trajectory contains a better box, but no test-time rule can select it. GRPO
  (right) only narrows the damage from iterating.}
  \label{fig:teaser}
\end{figure*}

\section{Related Work}
\label{sec:related}

\paragraph{Referring expression comprehension.}
Localizing an object from a natural-language description is a long-standing
task~\cite{refcocog,refl4}, addressed by specialist detectors~\cite{mdetr,groundingdino}
and, increasingly, by generative VLMs that emit box coordinates as
text~\cite{shikra,kosmos2,ferret,qwen2vl,qwen3vl}. Modern open VLMs are strong single-shot
grounders, which is precisely the regime our study targets: the question is not whether a
VLM can localize, but whether \emph{iterating} on its own localization helps.

\paragraph{Test-time reasoning and thinking with images.}
Chain-of-thought prompting~\cite{cot} and RL-trained reasoners such as
DeepSeek-R1~\cite{deepseekr1} show that spending more compute at inference improves
reasoning. A parallel line brings this to vision by reasoning \emph{through} visual
operations: CogCoM~\cite{qi2025cogcom} chains visual manipulations, GRIT~\cite{fan2025grit}
interleaves text with box-grounded references, ViGoRL~\cite{sarch2025vigorl} zooms into
predicted regions via multi-turn RL, RRVF~\cite{chen2025rrvf} closes a
reasoning--rendering--feedback loop, and Visual Planning~\cite{visualplanning} plans purely
in image space. \method{} is an instance of this line of work specialized to
\emph{self-correction of the same localization}: the model re-examines a rendered version of
its own prediction rather than gathering new information. Our contribution is not another
such method but a critical evaluation of whether this approach delivers \emph{deployable}
gains for grounding.

\paragraph{Self-correction and its limits.}
Self-Refine~\cite{selfrefine} and Reflexion~\cite{reflexion} iterate on a model's own
outputs, but Huang \etal~\cite{huangcorrect} show LLMs cannot reliably self-correct
reasoning without external feedback, and Stechly \etal~\cite{stechly2024selfverification}
identify self-\emph{verification} as the core limitation: models that can propose revisions
cannot reliably judge them. In vision, Liao \etal~\cite{liao2025can} study whether VLMs can
fix their own grounding errors, and Critic-V~\cite{zhang2025criticv} trains a separate critic
because the model's own feedback is unreliable. Our results are the spatial-grounding analog
and sharpen the diagnosis: even with rich \emph{visual} feedback and dedicated SFT$+$GRPO
training, the bottleneck is verification, and with the signals we test it leaves no
deployable test-time gain.

\paragraph{RL for grounding and verifiable rewards.}
GRPO~\cite{deepseekmath} removes the value function via group-relative advantages and suits
verifiable-reward tasks~\cite{ouyang2022training}. VLM-R1~\cite{vlmr1} shows a pure IoU
reward beats hand-designed composites for REC; UniVG-R1~\cite{bai2025univgr1},
Ground-R1~\cite{cao2025groundr1}, R1-VL~\cite{zhang2025r1vl}, and
Vision-R1~\cite{huang2026visionr1} apply R1-style training to multimodal grounding and
reasoning. We use GRPO with the simple IoU reward of VLM-R1, but find that in the
\emph{iterative} setting its benefit is to prevent degradation rather than to add accuracy
over single-shot prediction.

\paragraph{Evaluation integrity, calibration, and faithfulness.}
Best-of-$N$ and oracle-step reporting inflate apparent gains when the selection requires the
label, from pass@k~\cite{passk} to the oracle stopping behind reported self-correction
gains~\cite{huangcorrect}; our finding is a concrete case in spatial grounding, and our protocol
(\cref{sec:protocol}) is a drop-in remedy. Calibration of model confidence is a long-standing
concern~\cite{guo2017calibration}, and whether models ``know what they
know''~\cite{kadavath2022know} is precisely what a stopping rule needs. Work on explanation
faithfulness asks whether a model's stated rationale reflects its actual
computation~\cite{jacovi2020faithfully}; we give a spatial analog, asking whether a
``correction'' is aligned with an improvement in grounding, and find it is not. More
broadly, models often succeed for the wrong reasons~\cite{geirhos2020shortcut}; we show the
converse failure: an apparent reasoning gain that does not survive deployment-aware
measurement.

\section{A Competently Trained Iterative Grounder}
\label{sec:method}

\begin{figure*}[t]
    \centering
    \includegraphics[width=\textwidth]{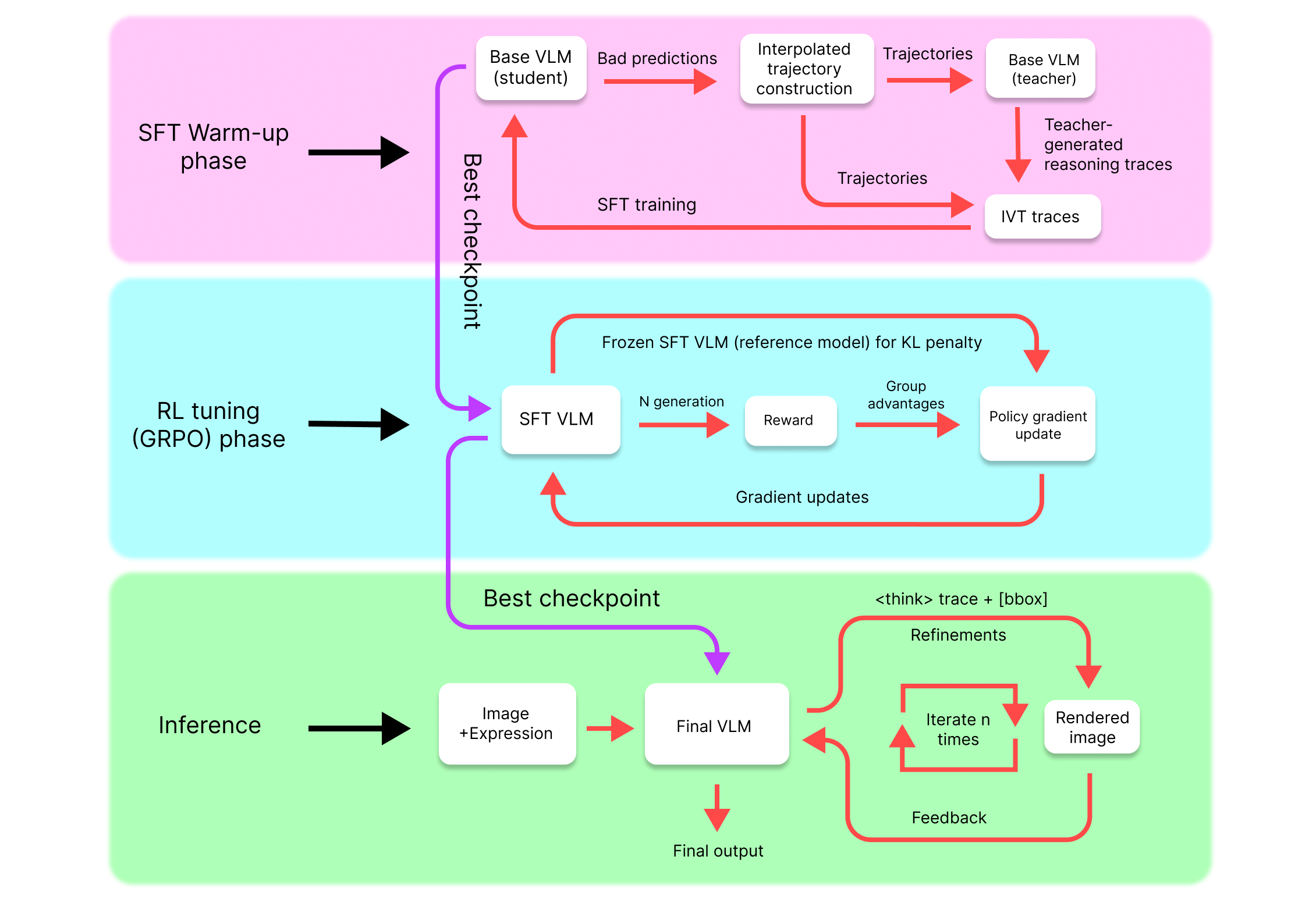}
    \caption{%
        \textbf{The \method{} recipe and inference procedure.}
        \emph{(Top)} SFT warm-up: the base model's own poor step-0 predictions seed box
        trajectories interpolated toward the ground truth; a teacher VLM writes
        step-conditioned corrective reasoning, and the student is fine-tuned on the
        resulting interleaved (text, rendered-image) traces via cross-entropy loss.
        \emph{(Middle)} GRPO: from the best SFT checkpoint we sample $N$ trajectories per
        prompt, score them by an IoU-based reward (Eq.~\ref{eq:reward}), and apply a
        group-relative policy-gradient update with a KL penalty to the frozen SFT reference.
        \emph{(Bottom)} Inference: the model emits a reasoning trace and box, observes its
        prediction rendered as a red overlay, and iteratively refines over $T$ steps within a
        single autoregressive continuation.
    }
    \label{fig:loop}
\end{figure*}

Our finding is a negative result, so it must target a competently trained system rather
than a strawman. We therefore study a faithful implementation of test-time visual
self-correction, \methodfull{} (\method{}), trained with the two-phase recipe that makes
the behavior emerge (\cref{fig:loop}). Given an image $I$ and a referring expression $e$,
the model generates a sequence of increasingly refined bounding boxes
$b^{(0)}, b^{(1)}, \ldots, b^{(T)}$ through visual feedback, rather than committing to a
single-pass answer.

\subsection{Iterative Visual Thinking}
\label{sec:ivt}

\paragraph{Step 0: Initial prediction.} The model receives the query image $I$ and
expression $e$ as a multimodal prompt. It generates a brief reasoning trace wrapped in
\texttt{<think>...</think>} tags, followed by a bounding box prediction $b^{(0)}$ in
normalized $[0, 1000]$ coordinates.

\paragraph{Steps $t = 1, \ldots, T$: Refinement.} Each refinement step has three stages:
\begin{enumerate}
    \item \textbf{Render.} The previous prediction $b^{(t-1)}$ is drawn as a
    semi-transparent red box overlay on the original query image $I$, producing a visual
    feedback image $I^{(t)}$.
    \item \textbf{Inject.} $I^{(t)}$ is appended to the model's context as a new visual input
    within the same assistant turn, followed by an explicit \emph{corrective instruction}
    directing the model to examine the red box, identify spatial errors, and output a
    corrected prediction. The model sees the original prompt, all prior reasoning, the
    rendered feedback, and this instruction.
    \item \textbf{Refine.} The model generates a new reasoning trace examining its previous
    prediction and outputs a refined box $b^{(t)}$.
\end{enumerate}

The entire sequence is generated within a \emph{single conversation turn} via
prefix-continuation: each step extends the same token sequence rather than starting a new
dialogue turn, preserving autoregressive coherence. We use $T{=}2$ refinement steps (three
predictions total). Coordinates are integers in $[0, 1000]$ normalized by image dimensions,
$x_\text{norm} = \lfloor x_\text{pixel} \cdot 1000 / W \rfloor$, decoupling the
representation from image resolution.

\subsection{Phase 1: SFT Warm-up}
\label{sec:sft}

Directly training iterative refinement with RL fails because the base VLM has no experience
with spatial self-correction: it produces nearly identical predictions at every step, and
GRPO cannot discover improvement behavior through random exploration alone (RL alone
degenerates to copying step~0). We address this with a supervised warm-up phase that teaches
the model \emph{what good refinement looks like}.

\paragraph{Student-prediction-based data synthesis.}
We use the student model's own predictions, rather than artificially perturbed ground-truth
boxes, as the starting point for trajectories. Our initial approach used random perturbations
of ground-truth boxes; we found empirically that this caused \emph{step-0 sandbagging} during
GRPO: the SFT-trained model learned to intentionally produce poor initial predictions
(mimicking the perturbed starting points), biasing RL rollouts and reward estimation.
Switching to student predictions resolved this, because the step-0 errors in SFT trajectories
then match the model's natural failure modes, exactly what GRPO encounters during training.
This approach exploits the base model's existing spatial grounding capability, which is strong
(${\approx}$80\% Acc@0.5) despite lacking self-correction ability, to generate realistic,
model-idiosyncratic errors. The pipeline has three phases:

\begin{enumerate}
    \item \textbf{Student predictions.} The base VLM generates single-pass predictions for
    each training sample, producing natural step-0 boxes $b^{(0)}_\text{student}$ with
    realistic errors.
    \item \textbf{IoU filtering.} We retain samples with student IoU in $[0.0, 0.85]$,
    discarding near-perfect predictions (no room to improve).
    \item \textbf{Trajectory + teacher reasoning.} A box trajectory is constructed from
    $b^{(0)}_\text{student}$ to $b_\text{GT}$ via linear interpolation. The teacher generates
    step-specific reasoning traces under a \emph{forward-reasoning} framing: at step 0 it sees
    the original image (no overlay) and writes as if reasoning about why the initial region was
    selected; at each subsequent step $t$ it sees the step $t{-}1$ rendered overlay, exactly
    matching what the student sees at inference, and writes 2--3 sentences examining the red box,
    identifying what is wrong (\eg, shifted, too large, wrong object), and explaining the spatial
    correction needed.
\end{enumerate}

The SFT dataset consists of examples with interleaved text (reasoning + coordinates), images
(rendered predictions), and corrective instructions. The corrective instructions, identical to
those injected at inference, are \emph{masked from loss} (labels set to $-100$), so the model
learns to generate reasoning and predictions but not the system-injected prompts. We train with
cross-entropy loss on the remaining assistant tokens using LoRA adapters.

\subsection{Phase 2: GRPO Training}
\label{sec:grpo}

Starting from the SFT-initialized model, we apply Group Relative Policy
Optimization~\cite{deepseekmath,deepseekr1} to optimize trajectory-level outcomes.

\paragraph{GRPO formulation.} For each training sample, we generate $N{=}8$ complete
trajectories from the current policy $\pi_\theta$. Advantages are computed relative to the
group mean:
\begin{equation}
    \hat{A}_i = \frac{R(\tau_i) - \mu_R}{\sigma_R + \epsilon}.
\end{equation}
The policy is updated via the REINFORCE objective with a KL penalty against the SFT reference
policy $\pi_\text{ref}$ and an entropy bonus:
\begin{equation}
    \mathcal{L} = -\mathbb{E}\!\left[\log \pi_\theta(a|s) \cdot \hat{A}\right] + \beta \, \overline{|\log \pi_\theta - \log \pi_\text{ref}|} - \lambda \, H(\pi_\theta),
\end{equation}
where $\beta = 0.04$ is the KL coefficient and $\lambda = 0.01$ the entropy bonus weight. The
KL term uses the mean absolute log-probability difference over generated tokens to regularize
against the frozen reference policy.

\subsection{Trajectory Reward}
\label{sec:reward}

Inspired by VLM-R1~\cite{vlmr1}, we adopt a deliberately simple reward that provides a clean
gradient signal for GRPO:
\begin{equation}
    R(\tau) = \underbrace{\text{IoU}(b^{(T)}, b_\text{GT})}_{\text{final-step IoU}} \;+\; 0.1 \cdot \underbrace{\frac{1}{T{+}1} \sum_{t=0}^{T} \mathbb{1}[\text{valid}(b^{(t)})]}_{\text{format fraction}},
    \label{eq:reward}
\end{equation}
where $b^{(T)}$ is the final-step prediction and $\mathbb{1}[\text{valid}(\cdot)]$ indicates a
parseable box. The $0.1$-scaled format term encourages well-formed predictions. In early
experiments a six-component reward (improvement, convergence, efficiency, format, regression,
and stagnation terms; 15 tunable parameters) produced unstable GRPO training: reward variance
across groups stayed high and per-component contributions partially cancelled, yielding noisy
gradients. The simple reward stabilized training, consistent with VLM-R1~\cite{vlmr1}.

\paragraph{Setup.}
The base model is Qwen3-VL-4B-Instruct~\cite{qwen3vl} with LoRA~\cite{lora} (rank~64,
$\alpha{=}128$) and 4-bit NF4 quantization~\cite{dettmers2023qlora}, trained on a single GPU.
Data is a balanced 1:1:1 mix of RefCOCOg~\cite{refcocog}, Ref-Adv-S~\cite{dong2026refadv}, and
Ref-L4~\cite{refl4} ($2{,}400$ train / $510$ val / $505$ test); images are processed at native
resolution. We report Acc@0.5/0.7/0.9 and mean IoU. For every configuration we
evaluate the \emph{full trajectory} and record per-step boxes, so we can separate oracle
best-step accuracy from accuracy under deployable inference policies, the central distinction
in this paper.

\section{The Self-Correction Mirage}
\label{sec:mirage}

\subsection{What the standard report shows}

\Cref{tab:main} reports the metrics the way best-of-$N$ predictors are typically scored, and the
way our pipeline scored them: for each test sample, the trajectory's \emph{best} box (highest
IoU against the ground truth) is kept. Read this way, the result is clean. Na\"ively iterating collapses the
base model by $31$pp; SFT warm-up repairs the collapse and surpasses the base on every metric
($+2.4$pp Acc@0.5); adding GRPO matches SFT on Acc@0.5 and slightly trails on the stricter
thresholds. One would conclude that test-time visual self-correction works once the model is
trained for it.

\begin{table*}[t]
  \centering
  \caption{\textbf{Standard (oracle best-step) report} on the mixed 3-way test set ($505$
  samples). Each entry scores the best box \emph{per sample} across the trajectory. Read this
  way, SFT$+$\method{} appears to beat the base by $+2.4$pp Acc@0.5. \Cref{tab:policies} shows
  this gain is unavailable in deployment.}
  \label{tab:main}
  \begin{tabular}{@{}lccccc@{}}
    \toprule
    Method & \method{} & Acc@0.5 & Acc@0.7 & Acc@0.9 & IoU \\
    \midrule
    Base (single-shot)        & \ding{55} & 0.796 & 0.709 & 0.455 & 0.719 \\
    Base $+$ \method{}        & \ding{51} & 0.487 & 0.442 & 0.283 & 0.442 \\
    \midrule
    SFT $+$ \method{}         & \ding{51} & \textbf{0.820} & \textbf{0.741} & 0.483 & \textbf{0.743} \\
    GRPO $+$ \method{}        & \ding{51} & \textbf{0.820} & 0.731 & \textbf{0.487} & 0.736 \\
    \bottomrule
  \end{tabular}
\end{table*}

\subsection{Removing the oracle}

The numbers in \cref{tab:main} require knowing, for each image, which step produced the best
box, information that itself requires the ground-truth box. No deployed system has this. Keeping
the best of several candidates by its match to the answer is oracle selection, equivalent to
pass@k~\cite{passk}, and it is the localization form of the oracle stopping that Huang
\etal~\cite{huangcorrect} traced behind reported self-correction gains. We therefore re-score
the \emph{same} trajectories under inference policies that select a step using only information
available at test time:
\begin{itemize}
  \item \textbf{single-shot}: take step~0 and never iterate;
  \item \textbf{last}: take the final step $b^{(T)}$;
  \item \textbf{last-valid}: take the last step that produced a parseable box;
  \item \textbf{stability~$\tau$}: stop once consecutive boxes overlap by IoU~$\ge\tau$;
  \item \textbf{movement~$\epsilon$}: stop once the box moves less than $\epsilon$ (normalized)
  between steps.
\end{itemize}
\Cref{tab:policies} reports the result: \emph{every deployable policy
loses to not iterating.} For the SFT model, single-shot ($0.796$) beats the best shippable
rule (movement, $0.685$) by $11.1$pp and the last-step by $15.6$pp; for GRPO, single-shot
($0.794$) beats the best rule ($0.756$) by $3.8$pp. The oracle row ($0.820$ for both) is
unreachable: it is exactly the gap between what the model \emph{could} achieve if it knew when
to stop and what it \emph{can} achieve in practice.

\paragraph{Even the ``collapse'' was an oracle number.}
The artifact reaches back to the headline phenomenon. The widely quoted $48.7\%$ for na\"ive
Base$+$\method{} (a $31$pp drop) is itself the best-step (and last-valid) figure: it silently
keeps the best of three boxes and discards unparseable ones. The deployable number is
the last-step accuracy a user would actually receive, $17.0\%$ (IoU $0.156$). The honestly measured
collapse is therefore $79.6\!\rightarrow\!17.0\%$, a \textbf{$63$pp} drop, larger than the
oracle-softened $31$pp usually quoted. Removing the oracle worsens this number rather than
improving it: without training, feeding a strong grounder its own rendered box roughly halves
its accuracy.

\begin{table*}[t]
  \centering
  \caption{\textbf{Oracle vs.\ deployable inference policies} on the $505$-sample test set
  ($T{=}2$). Only the oracle requires the ground truth. For both trained models, the best
  \emph{shippable} policy is \emph{single-shot} (do not iterate); every iterative policy is
  worse. Acc = Acc@0.5; \texttt{steps} is the mean number of forward passes used.}
  \label{tab:policies}
  \begin{tabular}{@{}lccc@{\hskip 1.2em}ccc@{}}
    \toprule
    & \multicolumn{3}{c}{SFT $+$ \method{}} & \multicolumn{3}{c}{GRPO $+$ \method{}} \\
    \cmidrule(r){2-4}\cmidrule(l){5-7}
    Policy & Acc & IoU & steps & Acc & IoU & steps \\
    \midrule
    single-shot (step~0)       & \textbf{0.796} & \textbf{0.716} & 1.00 & \textbf{0.794} & \textbf{0.714} & 1.00 \\
    last                       & 0.640 & 0.576 & 3.00 & 0.739 & 0.669 & 3.00 \\
    last-valid                 & 0.640 & 0.576 & 3.00 & 0.739 & 0.669 & 3.00 \\
    stability $\tau{=}0.90$     & 0.644 & 0.585 & 2.50 & 0.756 & 0.685 & 2.12 \\
    movement $\epsilon{=}0.05$  & 0.685 & 0.616 & 2.18 & 0.756 & 0.685 & 2.12 \\
    \midrule
    \emph{oracle (best step)}  & \emph{0.820} & \emph{0.743} & 1.35 & \emph{0.820} & \emph{0.736} & 1.14 \\
    \bottomrule
  \end{tabular}
\end{table*}

\subsection{GRPO is defensive, not additive}

\Cref{tab:policies} also reframes GRPO's contribution. The original framing credits GRPO with
``stabilizing refinement.'' That is accurate but limited: GRPO shrinks the gap between the
last-step and single-shot policies (from $-15.6$pp under SFT to $-5.5$pp) and reduces mean
per-step IoU degradation roughly $5\times$. But it never produces a policy that \emph{beats}
single-shot. GRPO makes iteration \emph{less harmful}; it does not make it \emph{helpful}. In
other words, the best thing GRPO teaches the model is to change its prediction less, which is
another way of saying it learns not to iterate.

\subsection{Why no test-time signal recovers the oracle gain}
\label{sec:strata}

\begin{figure}[htb]
  \centering
  \includegraphics[width=0.68\linewidth]{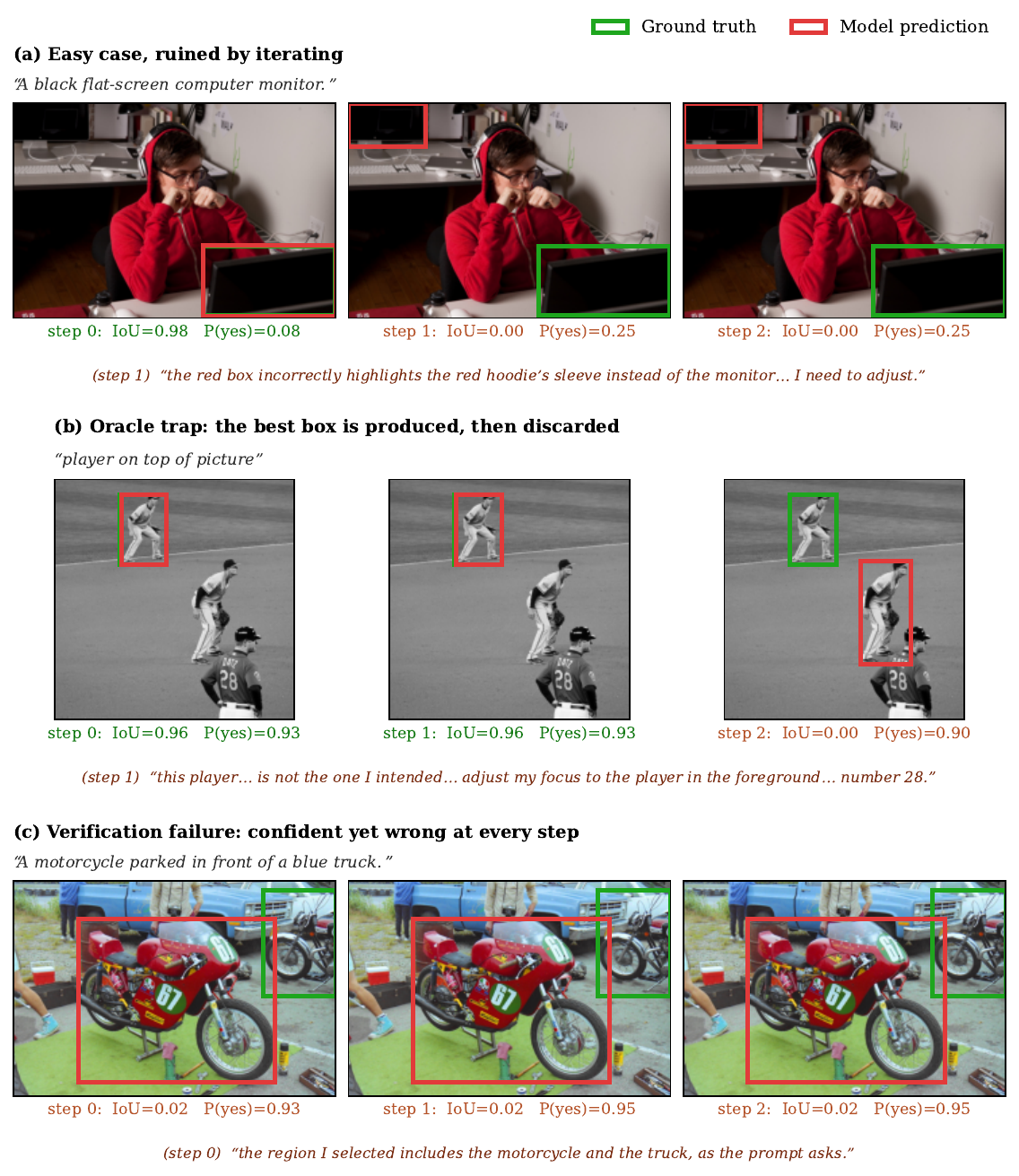}
  \caption{\textbf{Three failure modes of iterative visual self-correction}
  (GRPO\,$+$\,\method{}; \textcolor[HTML]{1ea61e}{green}~= ground truth,
  \textcolor[HTML]{e23a3a}{red}~= model prediction). \textbf{(a)}~An easy case the base model
  already solves (step~0 IoU $0.98$) is \emph{ruined} by iterating: the box drifts off the
  monitor by step~2. \textbf{(b)}~The \emph{oracle trap}: the model produces a near-perfect box
  at steps~0--1 (IoU $0.96$) but discards it at the last step (IoU $0.00$); only a label-aware
  oracle could keep the good box, and the model stays $90\%$ confident on the wrong one.
  \textbf{(c)}~A \emph{verification failure}: at every step the model confidently
  ($P(\text{yes})\!\approx\!0.95$) boxes the prominent foreground motorcycle rather than the one
  actually parked in front of the truck (the referent, in green); it cannot tell it is wrong.}
  \label{fig:qual}
\end{figure}

A clever adaptive policy could in principle approach the oracle if some observable signal told
it when a later step will help. \Cref{tab:strata} shows why none does. Stratifying by step-0
IoU, the oracle's gain is concentrated entirely in the $\sim$20\% \emph{hard} cases, where mean
IoU rises from $0.10$ to $0.20$ (an improvement that moves predictions from one failure regime
to another, rarely crossing the $0.5$ threshold). On the $66\%$ \emph{easy} cases iteration
almost always hurts ($63\%$ degrade under SFT; \cref{fig:qual}a,b). Step~0 is
already the best step for $76\%$ (SFT) / $90\%$ (GRPO) of samples, so the room to improve is
small to begin with. Within the hard cases, \emph{nothing observable predicts} which will
benefit: a later step helps only $39\%$ (SFT) / $14\%$ (GRPO) of the time, and the decision is
uncorrelated with expression length ($r\!\approx\!-0.14$) or box geometry. The signal a stopping
rule would need to exploit is, empirically, not in the inputs: it is in the ground truth.

\begin{table*}[t]
  \centering
  \caption{\textbf{Difficulty-stratified analysis} (GRPO and SFT, $505$ samples), by step-0
  IoU. The oracle's gain lives only in the hard stratum, moving failures to slightly-less-bad
  failures; easy cases mostly degrade. ``later helps/hurts'' = a refinement step changes IoU
  vs.\ step~0 by $>0.01$.}
  \label{tab:strata}
  \begin{tabular}{@{}llccccc@{}}
    \toprule
    Model & Stratum & \%\,samp. & IoU$_0$ & IoU$_{\text{oracle}}$ & later-helps & later-hurts \\
    \midrule
    \multirow{3}{*}{SFT}
      & Hard ($<0.5$)        & 20 & 0.104 & 0.198 & 39\% & 13\% \\
      & Med ($[0.5,0.75)$)   & 13 & 0.653 & 0.690 & 28\% & 55\% \\
      & Easy ($\ge0.75$)     & 66 & 0.917 & 0.921 &  9\% & 63\% \\
    \midrule
    \multirow{3}{*}{GRPO}
      & Hard ($<0.5$)        & 21 & 0.099 & 0.203 & 14\% &  0\% \\
      & Med ($[0.5,0.75)$)   & 13 & 0.655 & 0.657 &  7\% & 10\% \\
      & Easy ($\ge0.75$)     & 66 & 0.918 & 0.918 &  2\% &  8\% \\
    \bottomrule
  \end{tabular}
\end{table*}

\section{The Verification Bottleneck}
\label{sec:verification}

\Cref{sec:strata} shows that no \emph{input-derived} signal predicts when to stop. The most
direct remaining signal is the model's own judgment: ask it. Slow-thinking systems that
succeed in other domains do so partly because they can \emph{verify} candidate
solutions~\cite{stechly2024selfverification}; if the model could verify its own boxes, an
adaptive policy could keep the step it endorses.

\paragraph{A self-verification probe.}
For each refinement step we render $b^{(t)}$ on the image and ask the trained model a direct
yes/no question, ``is the red box a correct localization of the expression?'', and read
$P(\text{yes})$ from the logits~\cite{kadavath2022know}. We then use this confidence two ways:
\textbf{gated}, stop at the first step with $P(\text{yes})\!\ge\!\theta$; and \textbf{argmax},
take the step with the highest $P(\text{yes})$.

\paragraph{Confidence does not track correctness.}
\Cref{tab:verify} reports the outcome. Confidence is only weakly correlated with actual step
IoU (Pearson $r\!=\!0.229$ for SFT, $0.223$ for GRPO), and the model is poorly calibrated: mean
$P(\text{yes})\!\approx\!0.40$ even though most boxes are correct, \ie, it disbelieves its own
correct predictions (and, conversely, is confidently wrong; \cref{fig:qual}c). Consequently
every confidence policy \emph{still loses to not iterating}: SFT argmax reaches $0.723$ vs.\
single-shot $0.796$; GRPO argmax reaches $0.762$ vs.\ single-shot $0.794$. Tightening the gate
only makes things worse by forcing more iteration. For GRPO,
$\argmax P(\text{yes})$ picks the oracle step $84.6\%$ of the time yet \emph{still}
underperforms single-shot, because the oracle advantage itself is tiny (\cref{tab:policies})
and the $15\%$ of wrong picks cost more than the right picks gain. Verification would have to be
both accurate \emph{and} the gain would have to be large; neither holds.

\begin{table}[t]
  \centering
  \caption{\textbf{Self-verification as a stopping signal} ($505$ samples). $P(\text{yes})$ is
  the model's confidence that its rendered box is correct. Every confidence-gated or argmax
  policy underperforms \emph{single-shot}. Acc = Acc@0.5.}
  \label{tab:verify}
  \begin{tabular}{@{}lcc@{\hskip 1.5em}cc@{}}
    \toprule
    & \multicolumn{2}{c}{SFT $+$ \method{}} & \multicolumn{2}{c}{GRPO $+$ \method{}} \\
    \cmidrule(r){2-3}\cmidrule(l){4-5}
    Policy & Acc & IoU & Acc & IoU \\
    \midrule
    single-shot (step~0)            & \textbf{0.796} & \textbf{0.716} & \textbf{0.794} & \textbf{0.714} \\
    gated $P(\text{yes})\!\ge\!0.5$ & 0.693 & 0.626 & 0.756 & 0.682 \\
    gated $P(\text{yes})\!\ge\!0.9$ & 0.667 & 0.605 & 0.747 & 0.675 \\
    $\argmax P(\text{yes})$         & 0.723 & 0.650 & 0.762 & 0.687 \\
    \midrule
    \emph{oracle (best step)}       & \emph{0.820} & \emph{0.743} & \emph{0.820} & \emph{0.736} \\
    \midrule
    \multicolumn{5}{l}{\footnotesize $r(P(\text{yes}),\text{IoU})$: $0.229$ (SFT), $0.223$ (GRPO); mean $P(\text{yes})\!\approx\!0.40$.}\\
    \bottomrule
  \end{tabular}
\end{table}

\paragraph{Even a perfect external hint is not used.}
Self-verification asks the model to grade its own box; a sharper test hands it the answer.
Holding the step-0 text fixed, we change \emph{only} the box rendered on the fed-back image
before one refinement step: its own $b^{(0)}$ (control), the ground-truth box (a perfect hint),
or the GT box mirrored through the image centre (a wrong hint). \Cref{tab:counterfactual} shows
the loop is insensitive to the overlay's \emph{content}. Painting the correct box on the image
barely moves the prediction ($\text{IoU}(b^{(1)},b_{\text{GT}})$: $0.63\!\to\!0.64$ SFT,
$0.67\!\to\!0.68$ GRPO; it rescues only $14$--$18\%$ of step-0 misses, near the $11$--$13\%$
rescued by the control), and the wrong box is not tracked (overlap $\approx\!0.08$) yet still
degrades the answer by $8.1$/$5.6$\,pp. The model neither exploits a perfect external signal nor
rejects a corrupt one: it refines from the language prior and its own previous box. This
sharpens the diagnosis: the loop does not merely fail to \emph{verify} its boxes, it does not
\emph{read} the rendered evidence in the first place. GRPO is far more box-stable than SFT
(it leaves $b^{(0)}$ essentially unchanged under the wrong overlay in $79\%$ of samples vs.\
$36\%$), consistent with its defensive, low-iteration behavior.

\begin{table}[t]
  \centering
  \caption{\textbf{Counterfactual overlay} ($505$ samples, one refinement step). Only the box
  drawn on the fed-back image changes; the model's step-0 text is held fixed.
  $\text{IoU}(b^{(1)},b_{\text{GT}})$ measures whether the overlay helps or hurts the answer;
  $\text{IoU}(b^{(1)},\text{overlay})$ measures whether the refined box tracks the planted box.
  A perfect GT overlay barely helps; a wrong overlay is not tracked yet still hurts. (For the GT
  row the overlay \emph{is} the ground truth, so the two groups coincide.) Step-0 baseline IoU
  $0.72$/$0.71$.}
  \label{tab:counterfactual}
  \begin{tabular}{@{}lcc@{\hskip 1.5em}cc@{}}
    \toprule
    & \multicolumn{2}{c}{$\text{IoU}(b^{(1)},b_{\text{GT}})$} & \multicolumn{2}{c}{$\text{IoU}(b^{(1)},\text{overlay})$} \\
    \cmidrule(r){2-3}\cmidrule(l){4-5}
    Box drawn at step~1 & SFT & GRPO & SFT & GRPO \\
    \midrule
    own (its own $b^{(0)}$) & 0.63 & 0.67 & 0.80 & 0.89 \\
    GT (perfect hint)       & 0.64 & 0.68 & 0.64 & 0.68 \\
    wrong (mirrored)        & 0.55 & 0.61 & 0.09 & 0.08 \\
    \bottomrule
  \end{tabular}
\end{table}

\paragraph{Implication for slow thinking in grounding.}
Across four independent angles, box-geometry heuristics (\cref{tab:policies}), observable
difficulty (\cref{tab:strata}), self-verification confidence (\cref{tab:verify}), and a
counterfactual overlay that even hands the model the answer (\cref{tab:counterfactual}), no
deployable signal recovers the oracle, and the loop does not respond to the overlaid evidence
at all. The trained model can \emph{generate} a better box somewhere in its trajectory but
cannot \emph{identify} it. This is the verification gap, and it is exactly what
lets test-time reasoning pay off in domains with checkable
answers~\cite{stechly2024selfverification}. For spatial grounding the answer is not checkable
by the model itself, so additional test-time steps add cost without shippable benefit. We read
this not as a property of one model but as a direction: \emph{progress on iterative visual
grounding hinges on a trustworthy verifier}, whether a calibrated self-assessment, an external
critic~\cite{zhang2025criticv}, or a reward model that scores grounding faithfulness, rather
than on more or better refinement steps.

\section{An Honest-Trajectory Evaluation Protocol}
\label{sec:protocol}

The mirage in \cref{sec:mirage} is not a property of one model; it is a property of how
iterative grounders are scored. Best-step-per-sample reporting credits the model with a
selection it did not make. We propose a small, drop-in protocol that any iterative-grounding
paper can adopt so that a claimed self-correction gain is counted only when it is
\emph{attainable}.

\paragraph{Definitions.}
For a trajectory $b^{(0)},\dots,b^{(T)}$ with ground truth $b_\text{GT}$ and a metric $m$
(\eg, Acc@0.5), define:
\begin{itemize}
  \item \textbf{Oracle} $m_\text{or} = \mathbb{E}_x\big[\max_t m(b^{(t)},b_\text{GT})\big]$:
  the usual best-step number (a label-aware upper bound).
  \item \textbf{Shippable} $m_\pi = \mathbb{E}_x\big[m(b^{(\pi(x))},b_\text{GT})\big]$ for a
  \emph{label-free} stopping policy $\pi$ (\eg, single-shot, stability, movement,
  self-verification). Report the \emph{best} $\pi$ over a fixed, pre-registered set, and always
  include single-shot ($\pi\equiv0$) as a baseline.
  \item \textbf{Oracle gap} $\Delta = m_\text{or} - \max_\pi m_\pi$: how much of the reported
  headroom is unattainable by any label-free policy. A faithful method has $\Delta\!\approx\!0$.
  \item \textbf{Honest self-correction gain}
  $g = \max_{\pi:\,\pi\not\equiv0} m_\pi - m_{\text{single-shot}}$: the gain a deployed user
  would actually see from \emph{iterating} rather than stopping at step~0. Iteration is
  justified only if $g>0$.
\end{itemize}

\paragraph{Reading our system through the protocol.}
For both trained models, $g<0$ (\cref{tab:policies}): no \emph{iterating} policy beats
single-shot ($g=-11.1$pp SFT, $-3.8$pp GRPO), so the honest self-correction gain is negative
despite a large oracle ($0.820$). Because single-shot is itself the best shippable policy, the
oracle gap collapses onto it: $\Delta=2.4$pp (SFT) and $2.6$pp (GRPO), \ie, the entire
advertised $+2.4$pp headline is exactly the part no deployable policy can reach. We recommend
that iterative-grounding results always state the triple
$(m_{\text{single-shot}},\,\max_\pi m_\pi,\,m_\text{or})$ and foreground $g$ and $\Delta$.
Reporting $g$ also reframes design effort: a method should be judged on its \emph{shippable}
curve, which rewards work on \emph{verification} (knowing when to stop) rather than on producing
ever-larger oracle ceilings.

\paragraph{GRPO under the protocol.}
The protocol clarifies what GRPO buys. It does not raise $g$ above zero, but it lifts $g$ from
$-11.1$ to $-3.8$pp and shrinks the cost of running the loop to its last step
(single-shot\,$-$\,last) from $15.6$ to $5.5$pp, roughly halving per-step degradation, \ie, it
makes iteration \emph{less harmful} (the trajectory wanders away from the answer less) without
making it beneficial. This is a meaningful but defensive effect that the standard report hides
and our protocol surfaces.

\paragraph{Why ``small-but-unreachable'' is the right reading.}
A natural objection is that even the oracle ceiling ($79.6\!\rightarrow\!82.0$) is modest, so
why dwell on it. The point cuts the other way: the gain a paper could ever claim is a tiny,
ground-truth-dependent $+2.4$pp, while \emph{actually} iterating costs up to $15.6$pp. A
small-but-unreachable ceiling is a stronger negative result than a large-but-vague one, and it
is precisely the positive framing (``trained self-correction beats the base by $+2.4$pp'') that
this measurement dissolves.

\section{Discussion and Conclusion}
\label{sec:conclusion}

We asked whether test-time visual self-correction helps VLM grounding and found that a
pipeline which appears to win ($+2.4$pp Acc@0.5) delivers no deployable gain once the oracle
best-step selection is removed: the best shippable policy is to \emph{not iterate}, GRPO only
makes iteration less harmful, and the oracle's small gain, concentrated in hard cases that
remain failures, is unrecoverable from any observable signal, including the model's own
self-verification confidence and a counterfactual overlay that hands it the answer. The binding
constraint is \emph{verification}: the model can refine but cannot tell which attempt is right,
and it does not even read the rendered evidence it is given. We therefore argue that
slow-thinking for spatial grounding will pay off only when paired with a trustworthy verifier,
whether a calibrated self-assessment, an external critic~\cite{zhang2025criticv}, or a faithfulness reward
model, and we contribute an honest-trajectory protocol so that future iterative ``gains'' are
credited only when attainable.

\paragraph{What would change the verdict.}
Our result is a statement about a binding constraint, not an impossibility proof. The natural
next step is to break the verification bottleneck directly: train a separate box verifier rather
than relying on the policy's own confidence, or add a faithfulness-scored reward that rewards
the model for endorsing only grounded boxes. A verifier that is both accurate \emph{and}
operating on a task with a larger attainable gain (more steps, harder distributions, or
multi-object settings where re-examination genuinely adds information) could make iteration pay.
The contribution here is to show that, absent such a verifier, more or better refinement steps do
not help, and to give the measurement that makes this visible.

\paragraph{Limitations.}
We study one model family (Qwen3-VL-4B) at modest training scale ($2{,}400$ samples, single GPU)
and a 2-step loop on single-object REC; larger models, more steps, or segmentation/multi-object
tasks may behave differently. Our self-verification probe uses a single prompt and the policy
model itself as verifier; a separately trained verifier could fare better and is the natural next
experiment. We do not claim iterative grounding is impossible; only that, under faithful
deployment-aware measurement, the gains reported by best-step selection do not materialize, and
the binding constraint is verification. These limitations bound our \emph{positive} claims; they
do not weaken the negative finding, which is a property of the evaluation protocol and reproduces
across both trained configurations.

{
    \small
    \bibliographystyle{ieeenat_fullname}
    \bibliography{references}
}

\clearpage
\appendix
\section*{Appendix}
\section{Full Stopping-Rule Sweep}
\label{sec:suppl_sweep}

\Cref{tab:full_sweep} reports every inference policy we evaluated, for all three
configurations (Base, SFT, GRPO) under the \method{} loop, on the $505$-sample mixed test set.
This is the complete data behind \cref{tab:policies} and the headline numbers in
\cref{sec:mirage}. The pattern is uniform: for every \emph{trained} configuration the best
deployable policy is single-shot (step~0), and the oracle row, reachable only with the
ground-truth box, sits above all of them. For the Base configuration, the only policies that
exceed single-shot are themselves label-aware (last-valid coincides with the oracle here because
the base model frequently emits an unparseable box at the last step, and discarding it is a
soft oracle). One row deserves clarification: the Base ``single-shot (step~0)'' entry
($0.414$ Acc@0.5) is the base model's first prediction \emph{under the \method{} prompt}, and
is well below the base model's native single-shot accuracy ($0.796$, \cref{tab:main}, measured
with the plain grounding prompt). The multi-step prompt format alone degrades the untrained
model's first output; for the trained models, step~0 under \method{} recovers to the native
single-shot level ($0.796$ SFT, $0.794$ GRPO).

\begin{table}[H]
  \centering
  \caption{\textbf{Complete inference-policy sweep} ($505$ samples, $T{=}2$). ``steps'' is the
  mean number of forward passes. Only the oracle (and, for Base, last-valid) requires the
  ground truth. The true na\"ive collapse is the Base \emph{last} row: $0.170$ Acc@0.5.}
  \label{tab:full_sweep}
  \begin{tabular}{@{}llccccc@{}}
    \toprule
    Config & Policy & IoU & Acc@0.5 & Acc@0.7 & Acc@0.9 & steps \\
    \midrule
    \multirow{6}{*}{Base}
      & single-shot (step~0)     & 0.377 & 0.414 & 0.374 & 0.246 & 1.00 \\
      & last                     & 0.156 & 0.170 & 0.156 & 0.093 & 3.00 \\
      & last-valid               & 0.442 & 0.487 & 0.442 & 0.283 & 2.30 \\
      & stability $\tau{=}0.90$   & 0.380 & 0.418 & 0.378 & 0.248 & 1.54 \\
      & movement $\epsilon{=}0.05$& 0.380 & 0.418 & 0.378 & 0.248 & 1.54 \\
      & \emph{oracle}            & \emph{0.442} & \emph{0.487} & \emph{0.442} & \emph{0.283} & 1.10 \\
    \midrule
    \multirow{6}{*}{SFT}
      & single-shot (step~0)     & \textbf{0.716} & \textbf{0.796} & \textbf{0.711} & \textbf{0.454} & 1.00 \\
      & last                     & 0.576 & 0.640 & 0.477 & 0.192 & 3.00 \\
      & last-valid               & 0.576 & 0.640 & 0.477 & 0.192 & 3.00 \\
      & stability $\tau{=}0.90$   & 0.585 & 0.644 & 0.487 & 0.261 & 2.50 \\
      & movement $\epsilon{=}0.05$& 0.616 & 0.685 & 0.562 & 0.263 & 2.18 \\
      & \emph{oracle}            & \emph{0.743} & \emph{0.820} & \emph{0.741} & \emph{0.483} & 1.35 \\
    \midrule
    \multirow{6}{*}{GRPO}
      & single-shot (step~0)     & \textbf{0.714} & \textbf{0.794} & \textbf{0.709} & 0.469 & 1.00 \\
      & last                     & 0.669 & 0.739 & 0.657 & 0.432 & 3.00 \\
      & last-valid               & 0.669 & 0.739 & 0.657 & 0.432 & 3.00 \\
      & stability $\tau{=}0.90$   & 0.685 & 0.756 & 0.675 & 0.446 & 2.12 \\
      & movement $\epsilon{=}0.05$& 0.685 & 0.756 & 0.675 & 0.446 & 2.12 \\
      & \emph{oracle}            & \emph{0.736} & \emph{0.820} & \emph{0.731} & \emph{0.487} & 1.14 \\
    \bottomrule
  \end{tabular}
\end{table}

\section{Self-Verification Probe}
\label{sec:suppl_probe}

For the self-verification experiment (\cref{sec:verification}), at each step $t$ we render the
predicted box $b^{(t)}$ as a semi-transparent red overlay on the original image and append a
yes/no question of the form: \emph{``The red box marks a predicted location for: `\{expression\}'.
Is this box a correct localization? Answer yes or no.''} We read the next-token logits and
compute $P(\text{yes}) = \sigma(\ell_{\text{yes}} - \ell_{\text{no}})$ from the logits of the
``yes''/``no'' tokens. The \textbf{gated} policy stops at the first step whose $P(\text{yes})$
exceeds a threshold $\theta$ (we report $\theta\in\{0.5,0.9\}$); the \textbf{argmax} policy
returns the step with the highest $P(\text{yes})$ over the trajectory. Pearson correlations
between $P(\text{yes})$ and step IoU, and the mean $P(\text{yes})$, are reported in
\cref{tab:verify}.

\section{Counterfactual-Overlay Construction}
\label{sec:suppl_cf}

For the counterfactual experiment (\cref{tab:counterfactual}), we generate step~0 normally
(greedy) and then build the step-1 input three ways that differ \emph{only} in the box drawn on
the fed-back image, holding the model's step-0 text fixed: \textbf{own} (its own $b^{(0)}$, the
normal loop), \textbf{GT} (the ground-truth box, a perfect hint), and \textbf{wrong} (the
ground-truth box mirrored through the image centre). For each condition we record
$\text{IoU}(b^{(1)},b_{\text{GT}})$ (does the overlay help the answer) and
$\text{IoU}(b^{(1)},\text{overlay})$ (does the refined box track the planted box). As a sanity
check, the \textbf{own}-condition step-0 IoU reproduces the main-evaluation step-0 IoU exactly
for all $505$ samples in both models, confirming the overlay procedure is identical to the
evaluation pipeline.

\end{document}